\documentclass{article}

\usepackage{arxiv}

\usepackage[utf8]{inputenc} 
\usepackage[T1]{fontenc}    
\usepackage{hyperref}       
\usepackage{url}            
\usepackage{booktabs}       
\usepackage{amsfonts}       
\usepackage{nicefrac}       
\usepackage{microtype}      
\usepackage{lipsum}
\usepackage{graphicx} 
\graphicspath{ {./images/} }

\usepackage{natbib}

\usepackage{multirow}
 \usepackage{graphicx}
  \usepackage{comment}
  \usepackage{color}
  \usepackage{caption}
  \usepackage{subcaption}
  \usepackage[draft]{todonotes}
  \usepackage{listings}
  \usepackage{lscape}
  \usepackage{subfiles}
  
  \usepackage{algpseudocode}
\usepackage{algorithm}
  
  \usepackage{times,latexsym}
\usepackage{url}
\usepackage{geometry}

\usepackage{times}
\usepackage{latexsym}
\usepackage{amsmath}
\usepackage{multirow}
\usepackage{url}
\usepackage{xspace}
\usepackage{graphicx}
\usepackage{amsfonts}
\usepackage{mathtools,amssymb}
\usepackage{comment}
\usepackage{todonotes}
\usepackage{dblfloatfix}
\usepackage{microtype}
\usepackage{booktabs}

\usepackage{amsmath}

\newcommand{\projectNameIII}{HNMT\xspace}

\title{A Framework for Hierarchical Multilingual Machine Translation}

\author{
Ion Madrazo Azpiazu \\
  People and Information Research Team\\
  Dept. of Computer Science\\
 Boise State University\\
  Boise, ID \\
  \texttt{ionmadrazo@u.boisestate.edu} \\
   \And
Maria Soledad Pera \\
  People and Information Research Team\\
  Dept. of Computer Science\\
 Boise State University\\
  Boise, ID \\
  \texttt{solepera@boisestate.edu} \\
}

\begin{document}
\maketitle
\begin{abstract}
Multilingual machine translation has recently been in vogue given its potential for improving machine translation performance for low-resource languages via transfer learning. Empirical examinations demonstrating the success of existing multilingual machine translation strategies, however, are limited to experiments in specific language groups. In this paper, we present a hierarchical framework for building multilingual machine translation strategies that takes advantage of a typological language family tree for enabling transfer among similar languages while avoiding  the negative effects that result from incorporating languages that are too different to each other. Exhaustive experimentation on a dataset with 41 languages demonstrates the validity of the proposed framework, especially when it comes to improving the performance of low-resource languages via the use of typologically related families for which richer sets of resources are available.
\end{abstract}


\section{Introduction}

The explosive growth of text-based Web resources has been well-documented \cite{xu2018deep}. This, in theory, translates into a wealth of resources that are available to the masses. In practice, however, these resources are off-limit for populations that cannot read and understand the language in which each resource was originally written. Consequently, valuable resources written in popularly-spoken languages are out of reach for individuals who speak less-common ones, whereas resources written in minority languages rarely find their way into larger populations. Indeed, translators can fill this gap, but it is unfeasible for them to manually take ownership of this laborious and costly work on a large scale. This evidenced the need for machine translation: the task of taking a text in one language and translating it to another language in an automatic fashion \cite{sutskever2014sequence}. 



While as an area of study machine translation exists since the 1950s \cite{weaver1955translation}, it was not till the 1990s-early 2000s  when statistical approaches for machine translation showed prominence \cite{brown1990statistical,koehn2007moses}. It took some time for computing capabilities and text availability to converge into a new era of high-quality machine translation based on neural networks \cite{bahdanau2014neural,luong2015effective}. Nowadays, research pertaining to machine translation is categorized in three groups: rule based strategies that rely on hand written translation rules \citep{forcada2011apertium}, statistical techniques that learn rules based on parallel corpora \citep{koehn2007moses}, and neural machine translation strategies based on a encoder-decoder architecture \citep{sutskever2014sequence}, being the latter the most popular nowadays given its performance.

Most machine translation strategies build an encoder architecture that maps a text sequence in a source language to a vector representation and a decoder architecture that maps the vector representation to the same text sequence but in the target language, reducing the machine translation task to a purely bilingual task. 
The performance of this bilingual task is conditioned by two main factors: (1) quality/quantity of available parallel corpora, and (2) the similarity between the source and target language, defined in terms of the amount of linguistic patterns that both languages have in common. The more and better quality corpora it is available between the two languages, the better the translation quality it is expected to be. This is the reason why low-resource languages that tend to have less corpora available yield worse translation models. Additionally,  the more similar two languages are the better the translation will be. It is not the same to translate from Spanish to Portuguese, two languages that are closely related as they follow similar linguistic patterns, as it is to translate from English to Chinese, that share little to no similarity in lexical or grammatical patterns. 

Several approaches have been proposed in an attempt to address the two aforementioned problems, among which the translation via triangulation framework is the most prominent \citep{cohn2007machine,gollins2001improving}. In this framework, a translation is decomposed into multiple sub-translations in order to maximize corpora quality/quantity in each of the sub-translations in order to improve the performance of the final translation. As an example, instead of translating from Portuguese to Catalan, which might have reduced corpora available, a translation is first done from Portuguese to Spanish and then from Spanish to Catalan, improving the final translation performance given that both language pairs used (Portuguese-Spanish and Spanish-Catalan) have a considerable larger amount of parallel corpora available. This framework, however, has its own drawbacks, as a higher of sub-translations means a higher computational cost and a more prominent cumulative error (introduced at each sub-translation level). 

Multilingual machine translation, derived from multi-task training techniques, is a more recent framework that intends to address the corpora availability problem.  In this case the task of machine translation is no longer considered as a bilingual task, but as a multilingual task where multiple source and target languages can be simultaneously considered \citep{luong2015effective}. The objective of multilingual machine translation is to take advantage of knowledge in language pairs with large corpora availability and transfer it to lower resourced pairs by training them as part of the same model. For example, a single model can be trained to translate from Spanish to English and Catalan to English, with the expectancy that the performance of Catalan-English translations will get improved given that it has been trained together with a language with richer resources like  Spanish. Examples of multilingual machine translation models include those based on strategies that use a single encoder for all languages but multiple decoders \citep{dong2015multi}, or strategies that treat all languages as part of a single unified encoder-decoder structure \citep{ha2016toward}.



Even if existing multilingual machine translation strategies achieve language transfer to a degree, this transference only takes place when using specific language sets. Furthermore, these strategies ignore possible negative side-effects of including languages that are considerably different into a single model, i.e., training languages like Catalan and Spanish might be beneficial for performance, however, including a distant language like Chinese might decrease the overall performance of the same model. As a result, a state-of-the-art model such as the one described by \citet{ha2016toward} that includes all languages as part of a unified encoder-decoder structure would be sub-optimal when including language groups with strong differences. \citet{obj2} observed a similar behavior in the area of cross-lingual word embedding and concluded that putting all languages into a single space could act in detriment of the general model if it is not done in an organized fashion. 


Inspired by the idea of building a single model that can translate from multiple to multiple languages  \cite{ha2016toward} and the need of organization of languages when building multilingual strategies \cite{obj2}, we propose a Hierarchical Framework for Neural Machine Translation (\projectNameIII). \projectNameIII is a multilingual machine translation encoder-decoder framework that explicitly considers the inherent hierarchical structure in languages. For doing so, \projectNameIII exploits a typological language family tree, which is a hierarchical representation of languages organized by their linguistic similarity, in terms of grammar, vocabulary, and syntax, to name a few. In other words, \projectNameIII follows this natural connection among languages to encode and decode word sequences, in our case sentences. The hierarchical nature of languages allows \projectNameIII to only combine knowledge across languages with similar nature, while avoiding any negative knowledge transfer across distant languages. 



The main contributions of this work include:
\begin{itemize}
    \item A novel hierarchical encoder-decoder framework that can be applied to any of the popular state-of-the-at machine translation strategies to improve translation performance for low-resource languages.
    \item A comprehensive evaluation over 41 languages and 758 tasks to examine the extent to which language transfer is achieved.
    \item An analysis of the implications emerging from using the proposed framework for machine translation of low-resource languages.
\end{itemize}

\section{Related Work }
\label{sec:obj3:rw}
Machine translation techniques have been built using a variety of  strategies, including rule-based systems \citep{forcada2011apertium}, statistical machine translation \citep{koehn2007moses}, or neural machine translation strategies \citep{sutskever2014sequence}. In this work, we dedicate research efforts to neural machine translation strategies (NMT). More specifically, we focus on the enhancement of encoder-decoder strategies from a multilingual perspective. 
For this reason, we describe below related literature in the area of NMT and multilingual approaches for NMT. 
  
\textbf{Encoder-decoders strategies for NMT.} Encoder-decoder strategies were first proposed by \citet{sutskever2014sequence} as a solution for the inability of traditional neural networks for learning sequence-to-sequence mappings. This strategy was soon found lacking when translating long sentences given its need to compress all the sentence information into a low dimensional vector \citep{cho2014learning}. Several researchers tried to address this problem by allowing the decoder to have access to a larger amount of information, such as the previously generated word and the encoded sentence at any time step \citep{cho2014learning} or to the whole set of hidden states produced by the decoder via an attention mechanism \citep{bahdanau2014neural}. In order to obtain further training speed and translation quality, approaches presented later on tried remove the recurrent layers of the models, known to hinder  parallelization of models. With this purpose in mind, \citet{gehring2017convolutional} proposed a model based on Convolutional Neural Networks, while  \citet{vaswani2017attention} focused on just using layers purely based on attention. 

\textbf{Multilingual NMT.} Multilingual NMT strategies can be categorized by the degree to which they can share part of the architecture across different languages. \citet{dong2015multi} use a single encoder regardless of the language and rely on separate decoders for translation. \citet{luong2015effective} introduce a  strategy that uses one single encoder and decoder per language among all translation pairs.  \citet{firat2016multi} maintain the different encoders and decoders but share the attention mechanism across all translation pairs.  \citet{ha2016toward} propose to use one universal encoder and decoder that can handle any source and target language. This is achieved by providing the model with information of the language as an embedded parameter. 

Even if existing Multilingual NMT models can obtain varied ranges of transfer learning across languages, none of the  strategies we discussed takes advantage of the inherent hierarchical structure of the languages, that can beneficial to generate a more reliable language transfer among typologically similar languages, avoiding hindering the performance across distant languages.

\section{Method}
\begin{figure}[ht]
    \centering
    \caption{General bilingual machine translation architecture with 4 layers in both the encoder and the decoder.}
        \includegraphics[width=0.5\linewidth]{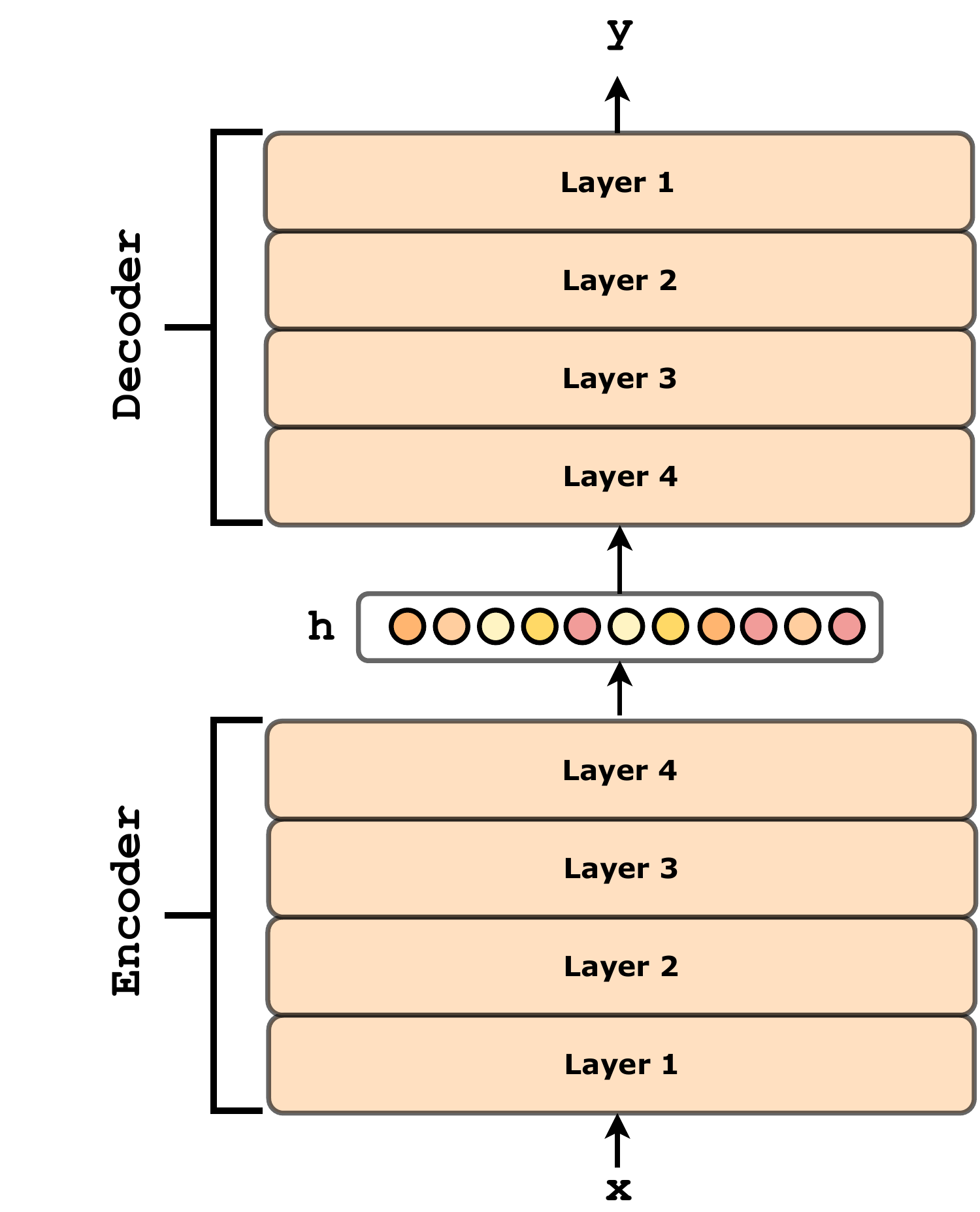}   
   
     \label{fig:obj3:singleMT}
    
\end{figure}

In this section, we describe the proposed framework for hierarchical multilingual machine translation, i.e.,  \projectNameIII. We first present a general sequence-to-sequence architecture used for neural machine translation, which we illustrate in Figure~\ref{fig:obj3:singleMT}. Then we explain how this general structure can be extended for multilingual machine translation. Lastly, we describe our proposed hierarchical framework illustrated in Figure~\ref{fig:obj3:generalArchitecture}.

\subsection{Neural Machine Translation}
State-of-the-art neural machine translation takes advantage of sequence-to-sequence models for translating a sequence $x$ (usually a sentence) in the source language $L_s$ to a sequence $y$ in the target language $L_t$. For doing so, the model is generally separated into an encoder module ($ENC_{L_s}$) capable for encoding $x$ into a vector representation $h$ of size $|h|$ and a decoder  ($DEC_{L_t}$) that aims at generating $y$ from $h$. 

Both the encoder and the decoder modules contain an equal amount of $N$ repeated layers 
that are used in a sequential way, as illustrated in Figure~\ref{fig:obj3:singleMT}. Starting from an input representation $x$ each encoding layer $ENC_{L_s}(i)$, is responsible for taking the representation $h_{{enc}_{i-1}}$ and generating $h_{{enc}_{i}}$, until it produces $h$. Once $h$ is generated each decoder layer $DEC_{L_s}(i)$, will take $h_{{dec}_{i}}$ and generate $h_{{dec}_{i-1}}$ until $y$ is generated. Following our naming convention: $h_{{enc}_{0}} = x $, $h_{{enc}_{N}} = h = h_{{dec}_{N}}$, and $h_{{dec}_{0}} = y $. If the model consists of 3 layers ($N=3$), the process of translating $x$ to $y$ requires $N*2=6$ steps:

\begin{equation}
\begin{split}
    x\xrightarrow{ENC_{L_s}(1)}h_{{enc}_{1}}\xrightarrow{ENC_{L_s}(2)}h_{{enc}_{2}}\xrightarrow{ENC_{L_s}(3)}h\\
    h\xrightarrow{DEC_{L_s}(3)}h_{{dec}_{2}}\xrightarrow{DEC_{L_s}(2)}h_{{dec}_{1}}\xrightarrow{DEC_{L_s}(1)}y
\end{split}
\end{equation}

Architectures for building each of the layers $ENC_{L_s}(i)$ and $DEC{L_s}(i)$ are manifold in the literature, being recurrent neural networks \cite{lstm}, convolutional neural networks \cite{gehring2017convolutional}, and transformers \cite{vaswani2017attention} the most widely accepted approaches. As previously stated, our proposed strategy is designed so that it can be applied to all of these architectures. However, for simplicity, we only showcase and discuss the practical application of \projectNameIII on a single architecture (see Section \ref{obj3sec:evalframework}). We use a Long Short Term Memory (LSTM) recurrent neural network \cite{lstm} given that it is an architecture with well-studied benefits and limitations. This enables us to isolate any phenomena introduced by this architecture from our framework so that we really focus our analysis on the advantages and disadvantages of our framework on its own.

\subsection{Multilingual Machine Translation}
In traditional (bilingual) neural machine translation, encoder and decoder modules are language specific, as captured by the subscripts in  $ENC_{L_s}$ and $DEC_{L_t}$. This means that an encoder for English ($ENC_{L_{en}}$) can be neither substituted by an encoder for Spanish ($ENC_{L_{es}}$) nor used to encode any $x$ that is not in English. Additionally, there is no guarantee that the representation $h$ is equivalent in any translation task, i.e., the representation $h$ that $ENC_{L_{en}}$ generates after being trained for English-Spanish translation is different from the representation generated by $ENC_{L_{en}}$ for English-Portuguese. Even inverse translation tasks, e.g., English-Spanish and Spanish-English, are considered to be separate tasks as there is no knowledge sharing across the two translations, resulting in different performance for each of the two. This poses a strong limitation to any language transfer in the translation task as every encoder/decoder is not only language specific but also task specific.

The goal of multilingual machine translation is indeed to address this limitation by generating models that can transfer language knowledge across tasks. This is achieved by frameworks from the multi-task learning area, such as jointly training two models for several tasks sharing part of the model weights \cite{firat2016multi,ha2016toward}. For example, for generating a model that can translate from both Spanish and Portuguese to English the model would be trained using pairs from both tasks, separate Spanish and Portuguese encoders but a single English decoder. This training strategy enables training the English decoder using data from both tasks (Spanish-English and Portuguese-English), benefiting from a larger aligned corpora and therefore achieving better decoding and translation performance. As described in Section~\ref{sec:obj3:rw}, different strategies have been proposed in literature for multilingual machine translation. However, to the best of our knowledge, all of them consider the encoders and decoders as a atomic unit that cannot be separated any further, just differentiating from each other by how many full decoders or encoders the model uses for multilingual translation purposes, i.e., one-to-many, many-to-one, or one-to-one \cite{firat2016multi,ha2016toward}. One of the limitations of these models is that they consider all languages to be of same nature, meaning that all languages are combined into a single encoder/decoder without any organization, ignoring the fact that some languages might indeed benefit each other while others would hinder the final performance of the translation task. This has demonstrated to be the case in related areas such as cross-lingual word embedding generation \cite{obj2}. Instead, \projectNameIII is capable of incorporating further divisions inside the encoder/decoder allowing a more fine grained possibilities for determining which languages should share weights or not as we describe in the following section.

\begin{figure}[ht]
    \centering
    \caption{Description of the general architecture of \projectNameIII}
        \includegraphics[width=0.8\linewidth]{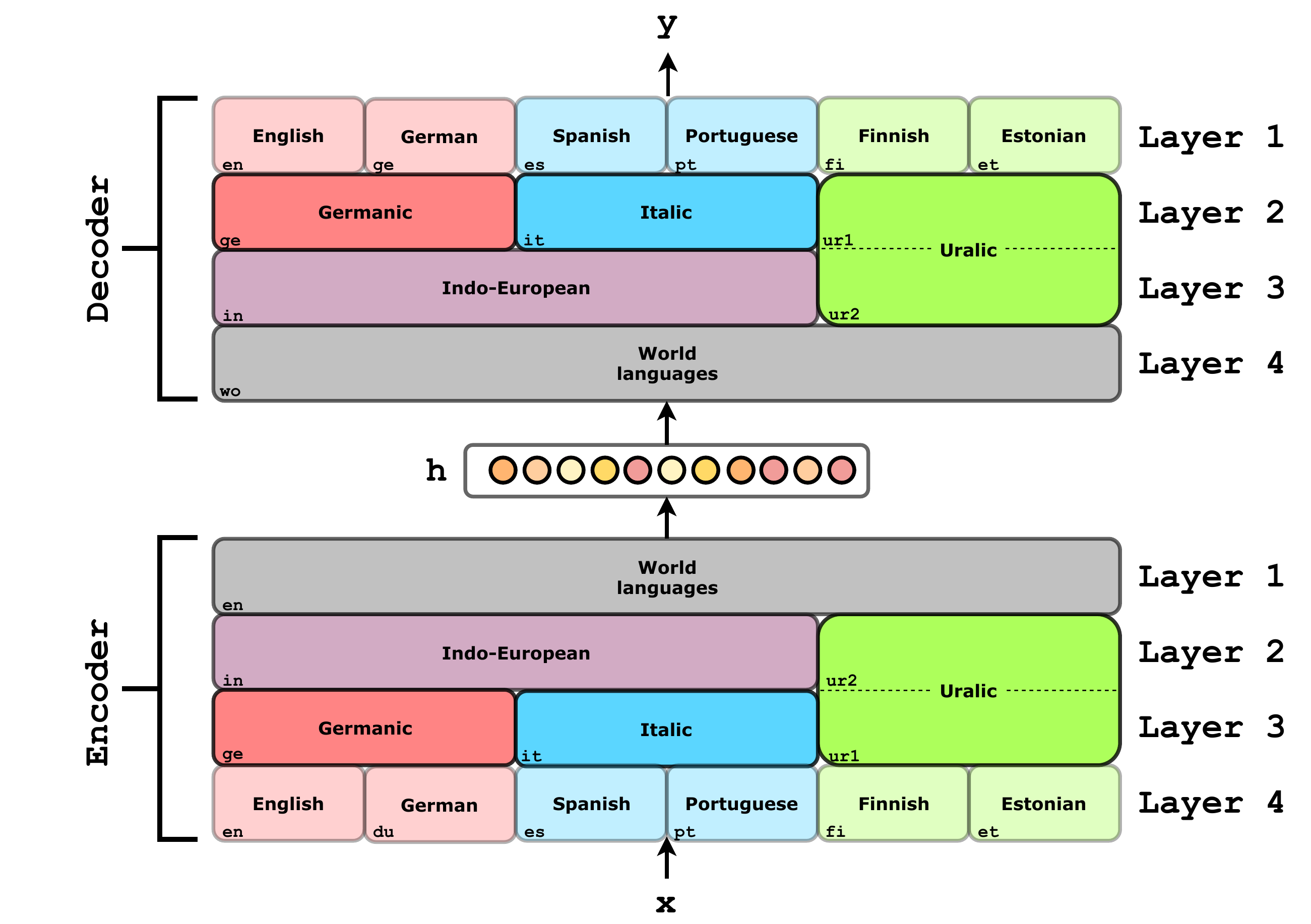}   
   
     \label{fig:obj3:generalArchitecture}
    
\end{figure}


\subsection{Hierarchical Multilingual Machine Translation}

For \projectNameIII, we define the multilingual machine translation task as a one-to-one task, meaning that \projectNameIII only contains a single encoder and a single decoder. However, each layer $ENC_{L_t}(i)$ or $DEC_{L_t}(i)$ is shared or not across languages depending on their similarity. This enables similar languages to share a larger amount of layers, fostering further language transfer among them, while different language share less layers, avoiding hindering the model's overall performance by forcefully combining language that are too distant. To delineate this inter-connectivity across languages \projectNameIII takes into account the hierarchical nature of languages by taking advantage of a typological language tree. As illustrated in Figure~\ref{fig:obj3:generalArchitecture} each family in the language tree corresponds to a layer of the encoder and the decoder. Each language always has a unique, non-shared-layer, which correspond to the very first layer of the encoder and very last layer in the decoder. This is due to the fact that we consider that each language to be different from any other even if it is to a small degree. This layer enables the model to capture these language-specific characteristics in both the encoder and the decoder. Additionally, \projectNameIII also incorporates one layer that is shared across all languages. This layer is located directly before and after the $h$ vector representation of the sentence, and its purpose is to unify how the model generates this vector regardless of the language used. The remaining intermediate layers are directly determined by the language tree.

For illustration purposes consider the translation task from Spanish to English versus the same task but from Spanish to Finnish. Based on the structure described in Figure~\ref{fig:obj3:generalArchitecture}, Spanish to English translating requires the following 8 steps:

\begin{equation}
\begin{split}
     x \xrightarrow{ENC_{es}} h_{{enc}_{1}} \xrightarrow{ENC_{it}} h_{{enc}_{2}}  \xrightarrow{ENC_{in}} h_{{enc}_{3}}  \xrightarrow{ENC_{wo}} h\\
    h \xrightarrow{DEC_{wo}} h_{{dec}_{1}} \xrightarrow{DEC_{in}} h_{{dec}_{2}}  \xrightarrow{DEC_{ge}} h_{{dec}_{3}}  \xrightarrow{DEC_{en}} y\\
\end{split}
\end{equation}
\noindent
where $ENC_{l}$ and $DEC_{l}$ refer to the encoder and decoder layers of language $l$ respectively. 

Spanish to Finnish translation, on the other hand, requires the following steps:
\begin{equation}
\begin{split}
     x \xrightarrow{ENC_{es}} h_{{enc}_{1}} \xrightarrow{ENC_{it}} h_{{enc}_{2}}  \xrightarrow{ENC_{in}} h_{{enc}_{3}}  \xrightarrow{ENC_{wo}} h\\
    h \xrightarrow{DEC_{wo}} h_{{dec}_{1}} \xrightarrow{DEC_{ur2}} h_{{dec}_{2}}  \xrightarrow{DEC_{ur1}} h_{{dec}_{3}}  \xrightarrow{DEC_{fi}} y\\
\end{split}
\end{equation}

It is important to note that, as reflected in  Figure~\ref{fig:obj3:generalArchitecture}, the language tree is not an equally balanced tree, meaning that the number of families from any language to the root node is different. In the example, the number of families from Spanish to the root is two, while the number of families from Finnish to the root is just one. This characteristic of the tree directly conflicts with the requirement of most existing sequence-to-sequence models to contain a same amount of layers in the encoder and the decoder. Additionally, some language might contain more families than layers are used in the model, i.e., if the layer number is chosen to be $N=3$ the families of English, German, Spanish, and Portuguese would not fit into the model. In order to address both of these concerns, we conduct a two-step preprocessing of the tree. First, we limit the tree to have $N-2$ layers, pruning any family that does not meet this constraint. 
Thereafter, we duplicate any leaf node that is not in the layer $N-2$ of the tree, e.g., in the sample tree in Figure~\ref{fig:obj3:generalArchitecture}, the Uralic family is duplicated to adhere to this constraint. 





\subsection{Training \projectNameIII for Sentence Translation}
\projectNameIII takes advantage of stochastic gradient descend for learning the weights of its model. Different from a traditional machine translation strategy, \projectNameIII utilizes multiple datasets with different languages for training. Training is conducted in a round-robin fashion with respect to the datasets, i.e., one epoch of each dataset is trained in a sequential manner. Each dataset epoch is divided into batches  of  sentence pairs $\beta_{batch}$, for which the loss function is computed, backtracked and parameters tuned. We use cross entropy as loss function and Adaptive Movement Estimation \citep{kingma2014adam} as optimizer with a learning rate of $\beta_{lr}$. The training will continue until no improvement is found on the training set for the average loss across all datasets in the last 10 epochs. In order to avoid overfitting, the model selected for testing is the model that achieves best performance in a separated validation set. Refer to Section~\ref{sec:obj3:hyper} for further details on how we split the datasets and tune hyper-parameters.








\section{Evaluation Framework}
\label{obj3sec:evalframework}
In this section, we describe the evaluation framework used for examining the performance of \projectNameIII and showcasing its advantages with respect to existing baselines. 

\subsection{Data}
We use the GlobalVoices parallel corpora \cite{globalvoices} for training and evaluation purposes. This dataset is comprised of bilingual corpora for most combinations across 41 languages, totaling 758 different tasks, i.e., pairs of languages for translation. Each task contains a varying amount of parallel sentence pairs that go from less than 10k sentences (in the case of Catalan-English) to more than half a million (in the case of Spanish-English). The strong variation of corpora available for each task mimics a real world scenario where few languages are very rich in resources while many barely have resources associated with them, making this dataset ideal for our experiments.

In order to input words to a neural machine translation model they first need to be converted into a numerical vector representation. For doing so, we take advantage of the cross-lingual word embeddings generated by \citet{obj2}, tailored to low-resource scenarios, a case we consider of specific interest in our experiments. 

Finally, we use the language tree described in \citet{ethnologue} on our experiments. This tree can sometimes be overly detailed, containing too many names describing nearly the same family of languages. For this reason, we prune the original tree to remove family names that can be treated as redundant for translation purposes. For example, having both Central Iberian and Castilian as family for the Spanish language is redundant. For pruning, we define the following criteria: Any family that contains exactly the same amount of languages as its parent is removed.

\subsection{Validation and Hyper-parameter Tuning}
\label{sec:obj3:hyper}

Each of the 758 task specific datasets considered in this study is randomly separated into 3 splits using  70\%, 10\%, and 20\% of the sentence pairs for training, validation, and testing, respectively. The training set used for learning the weights of the model. The validation portion is used for selecting the best model among the ones generated during training. Finally, the testing set is only used for measuring the performance of the final model. Disjoint from these 3 sets, we held-out a development set of 20k Spanish-English sentence pairs. This development set is only used for verifying the correctness of the implementation and tuning hyper-parameters. No sentence pair in this held-out set is ever included in any of the train, validation, or testing sets.

Hyper-parameters where manually selected, meaning that no exhaustive/automatic hyper-parameter tuning strategy was applied. The final hyper-parameters used in the experiments are:
$\beta_{batch}=768$,  $\beta_{lr}=0.01$, $|h|=512$, $N=5$, Layer-type$=$LSTM.  It is true that the number of weights selected for our strategy is comparably smaller than what most state-of-the-art strategies currently use\cite{johnson2017google}. In fact, for \projectNameIII we use $|h|=512$ and $N=5$, whereas for current strategies $|h|=1024$ and $N=8$ are customary. This was a compromise we had to take in order to balance for the large number of tasks we consider in the study, i.e., 758, compared with the \textit{less than a dozen} tasks most current studies consider \cite{artetxe2017unsupervised,johnson2017google}.

\subsection{Baselines}
To contextualize the performance of \projectNameIII, we compare its performance to that obtained by four baselines: a traditional bilingual baseline (Many-to-many) and three multilingual baselines (One-to-many, Many-to-one, One-to-one).

\begin{enumerate}
    \item \textbf{Many-to-many}. This model resembles the traditional bilingual machine translation strategy where each task has it own encoder a decoders.
    \item \textbf{One-to-many}. In this multilingual machine translation model one encoder is used regardless of the language and a different decoder for each language.
    \item \textbf{Many-to-one}. Opposite to the previous model this one uses a single decoder for all he language but multiple encoders, one per language.
    \item \textbf{One-to-one}. This is a universal machine translation model that utilizes just one encoder and one decoder for all the languages considered.
\end{enumerate}

Models that have a single decoder require some explicit information of the output language in order to enable the model to know the language in which it needs to generate the output sentence. For these models, we prepend the input of the model with a special token representing the language that needs to be generated similar to what is done by \citet{johnson2017google}. It is also important to note that unlike aforementioned models \projectNameIII does not require this token to operate, given that the last layer of the decoder is specific to the target language.


\subsection{Metric}
For measuring the performance of each task we take advantage of a traditional metric to the machine translation area: Bilingual Evaluation Under Study (BLEU) \cite{papineni2002bleu}.

\section{Results and Discussion}
\begin{figure}[ht]
    \centering
    \caption{BLEU score obtained for each of the bilingual tasks using a traditional machine translation baseline (many-to-many), organized by the number of sentence pairs available for the task. Results for tasks with more than 100k sentence are omitted for visualization purposes.}
        \includegraphics[width=0.8\linewidth]{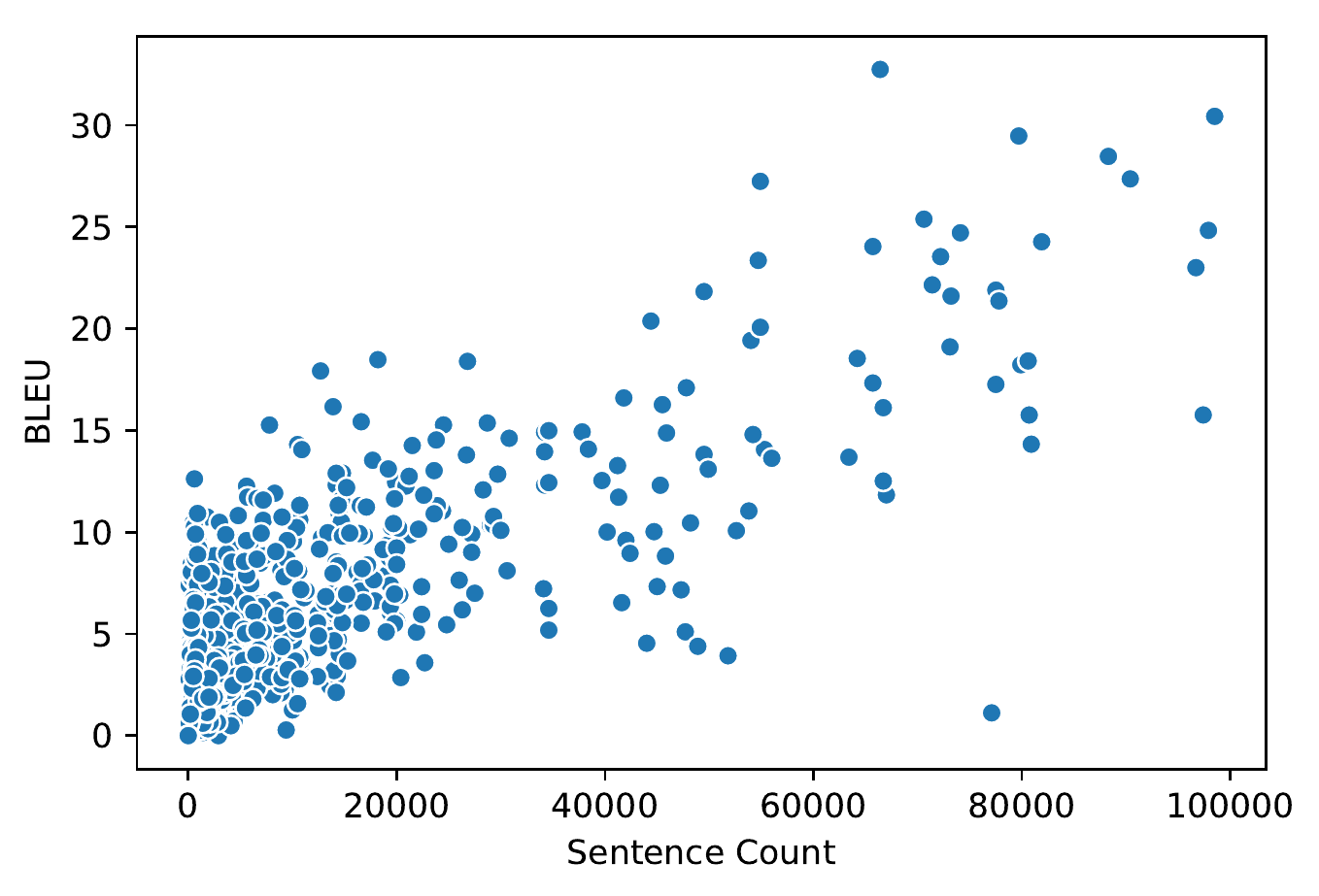}   
   
     \label{fig:obj3:scatterPlotSentenceCountM1}
    
\end{figure}

In order to fully analyze the performance of \projectNameIII, it is important to first understand what traditional (bilingual) machine translation strategies can achieve. Therefore, we start by analyzing the performance of the many-to-many model from different perspectives. For doing so, we train and test this model for each of the tasks defined by a language pair in the GlobalVoices dataset. In Figure~\ref{fig:obj3:scatterPlotSentenceCountM1}, we illustrate the performance of each task organized by the number of sentences available for the task. Emerging from the figure is a pattern that is inherent to the machine translation area: the more sentence pairs available, the better the performance of the model. This leads to an uneven scenario, one where strategies are  better performing, i.e., are more effective, for resource-rich languages than for low-resource ones.

\begin{figure}[]
    \centering
    \caption{BLEU score obtained for each of the bilingual tasks using a traditional machine translation baseline (many-to-many), grouped by the similarity between the source and target language for the task.}
        \includegraphics[width=0.8\linewidth]{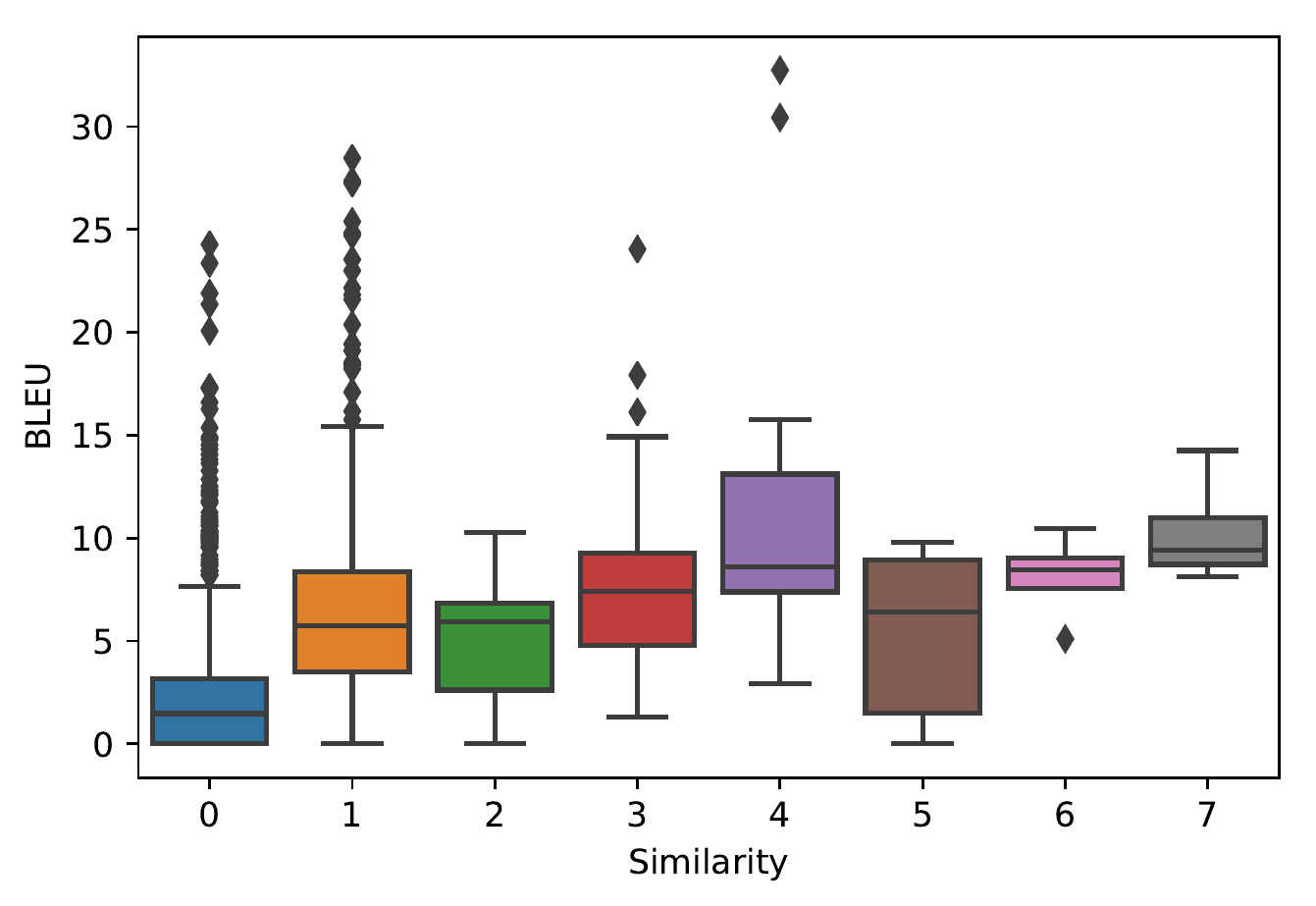}   
   
     \label{fig:obj3:boxplotSimilarityM1}
    
\end{figure}

Another issue affecting machine translation related to the direct connection that exists between performance and language similarity. While this is a fact that has been pointed out by several researchers \cite{cohn2007machine,gollins2001improving}, to the best of our knowledge, this has never been thoroughly studied. For demonstrating how this dependency behaves in our experiments, we define the similarity between two language as the number of parent family nodes they share. As an example, if we refer back to Figure~\ref{fig:obj3:generalArchitecture}, the similarity between English and German is 2 as they both have the Germanic and Indo-European families as parents, while the similarity between Finnish and English is 0 as they do not have any common parent family. We depict in Figure~\ref{fig:obj3:boxplotSimilarityM1} the performance of each of the tasks grouped by the similarity between the source and the target languages. From the figure, it is evident that the results follow a pattern where the more similar any two languages are, the better the quality of machine translation achieved for them.

These two patterns demonstrate (1) a real need in the area of machine translation for designing transfer learning strategy that can improve the performance of machine translation for low-resource languages,  and (2) a possibility to achieve valuable language transfer by using a proper organization of the languages, that makes is possible to take advantage of synergies among similar languages. These two patterns, further validate the premises that leaded us to design \projectNameIII.

In Table~\ref{tab:obj3:results} we present the results of 5 different machine translation models for each of the tasks in GlobalVoices dataset, grouped by source language. We include the traditional machine translation model (many-to-many), 3 frameworks for building multilingual machine translation that are representative of the current state-of-the-art, and our \projectNameIII framework. As shown in the table, our proposed strategy yields an average gain of 1.07 BLEU points over the traditional many-to-many strategy. This difference is statistically significant under paired T-test with a confidence interval of $p<0.05$. Largest improvements are found in languages such as Catalan, Portuguese, Italian, or Spanish, which we find not to be a coincidence but the result of \projectNameIII correctly integrating languages that share similarities. The lowest improvement is obtained for Oriya, an Indo-Aryan language that shares little similarity with respect to any of the remaining languages in the dataset. Among multilingual machine translation models (one-to-one, one-to-many, many-to-one), only the one-to-many model achieves an improvement over the traditional bilingual baseline. This is not a surprising result as, even if multilingual machine translation models have shown to improve over traditional machine translation with specific language combinations, they are known to under-perform when simultaneously dealing with either too many or too different languages \cite{johnson2017google}.

\begin{figure}[ht]
    \centering
    \caption{Distribution of BLEU scores yielded by \projectNameIII for individual translation tasks.}
    \includegraphics[width=0.8\linewidth]{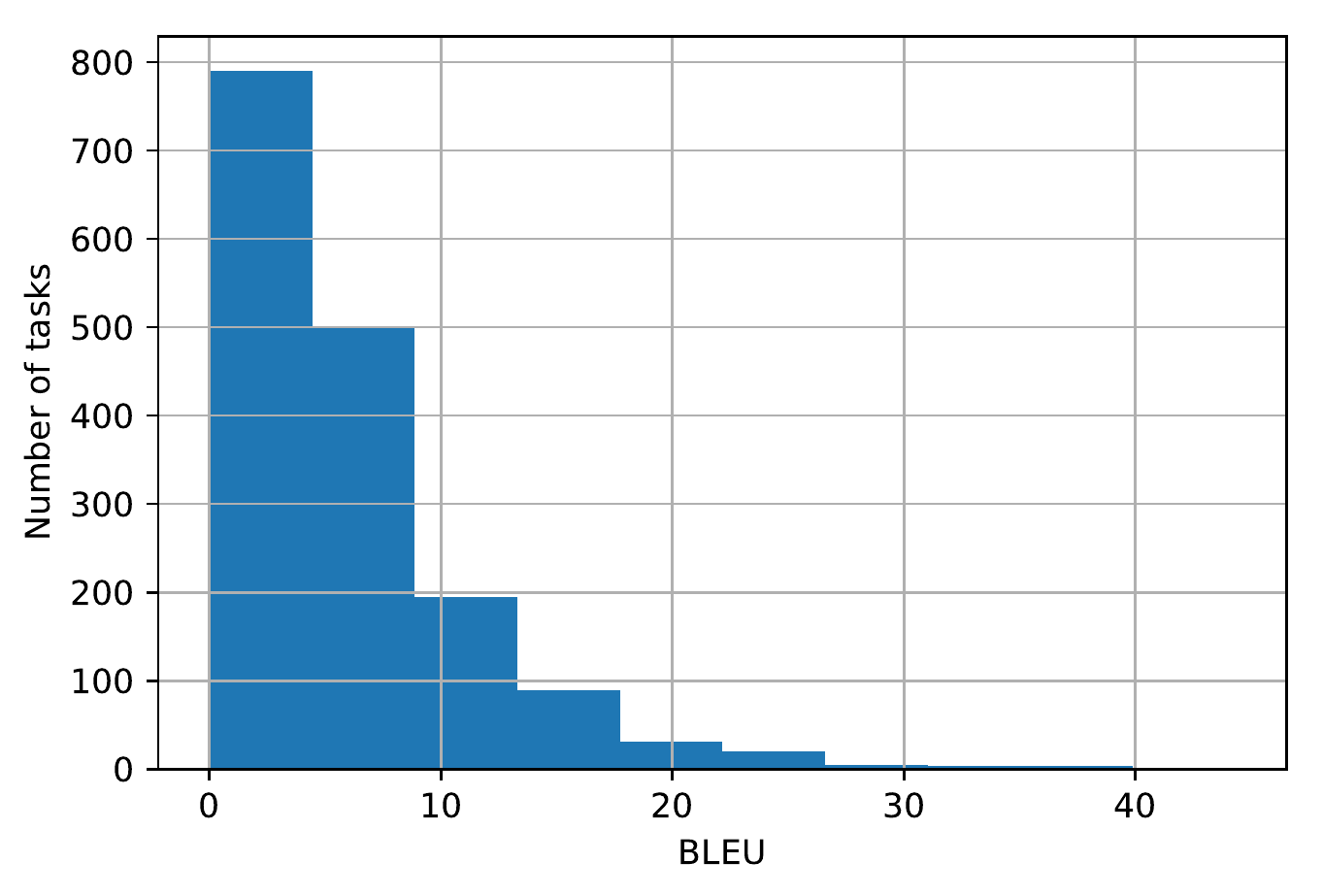}   
   \label{fig:obj3:histogramBLEU}
\end{figure}

Table~\ref{tab:obj3:results} captures average BLEU scores over all tasks that use the specified language as source. While average allow us to assess and compare performance across frameworks, it does not shine a light on translation pairs that greatly deviate from the average. To showcase the varied degrees of BLEU scores obtained by \projectNameIII for each of the 758 translation tasks, we included an histogram in  Figure~\ref{fig:obj3:histogramBLEU}. It can be appreciated in the figure that for most of the translation tasks performance ranges between 0 and 10 BLEU points. However, there are some cases for which BLUE is as high as 30-40. Not surprisingly, these cases align with popular, resource-rich languages like Spanish-English and Portuguese-English. At the opposite extreme, we see BLEU as low as 0.27. Once again, this is anticipated, as these low scores are the result of translation to/from low-resource (and often less recognized) languages, like Catalan to Oriya.

\begin{table}[h]
\centering
\caption{BLEU scores obtained by each of the baselines and the proposed model for each language considered in the study. Improvement denotes the difference with respect to the bilingual machine translation baseline (Many-to-many). Note that values shown for each language correspond to the average BLEU obtained using the language as source with respect to all other languages used as target. Language names are described using ISO-639 notation. }
\label{tab:obj3:results}
\resizebox{\textwidth}{!}{%
\begin{tabular}{@{}ccccccc@{}}
\toprule
& Many-to-many & One-to-many & Many-to-one & One-to-one & \projectNameIII & Improvement     \\ \midrule
am         & 1.51        & 1.65        & 1.32         & 1.15        & \textbf{2.02}  & 0.51        \\
ar         & 6.08        & 6.43        & 5.41         & 4.90        & \textbf{7.15}  & 1.06        \\
aym        & 4.08        & 4.28        & 3.43         & 3.04        & \textbf{5.07}  & 0.98        \\
bg         & 4.22        & 4.38        & 3.55         & 3.23        & \textbf{5.59}  & 1.36        \\
bn         & 8.92        & 9.11        & 7.85         & 7.53        & \textbf{10.28} & 1.36        \\
ca         & 5.62        & 5.96        & 4.73         & 4.43        & \textbf{7.33}  & 1.71        \\
cs         & 4.76        & 4.90        & 3.99         & 3.89        & \textbf{6.04}  & 1.28        \\
da         & 4.49        & 4.74        & 3.79         & 3.64        & \textbf{5.57}  & 1.08        \\
de         & 7.57        & 7.76        & 6.08         & 6.20        & \textbf{8.98}  & 1.42        \\
el         & 7.54        & 7.78        & 6.43         & 6.06        & \textbf{8.78}  & 1.25        \\
en         & 11.26       & 11.43       & 9.92         & 9.73        & \textbf{12.51} & 1.25        \\
eo         & 2.33        & 2.66        & 1.75         & 1.75        & \textbf{3.12}  & 0.78        \\
es         & 11.61       & 11.82       & 10.45        & 10.22       & \textbf{13.35} & 1.74        \\
fa         & 3.74        & 3.99        & 3.14         & 2.85        & \textbf{4.64}  & 0.90        \\
fil        & 2.60        & 2.91        & 2.09         & 1.97        & \textbf{3.31}  & 0.71        \\
fr         & 8.72        & 8.90        & 7.70         & 7.29        & \textbf{10.27} & 1.55        \\
he         & 1.21        & 1.36        & 0.92         & 0.95        & \textbf{1.65}  & 0.43        \\
hi         & 1.71        & 1.75        & 1.44         & 1.30        & \textbf{2.32}  & 0.61        \\
hu         & 4.50        & 4.79        & 3.61         & 3.40        & \textbf{5.46}  & 0.97        \\
id         & 4.05        & 4.21        & 3.04         & 3.02        & \textbf{4.87}  & 0.82        \\
it         & 7.84        & 8.04        & 6.73         & 6.47        & \textbf{9.35}  & 1.51        \\
jp         & 6.53        & 6.65        & 5.48         & 5.12        & \textbf{7.47}  & 0.95        \\
km         & 0.81        & 0.96        & 0.64         & 0.57        & \textbf{1.22}  & 0.41        \\
ko         & 3.67        & 4.05        & 2.99         & 2.79        & \textbf{4.58}  & 0.90        \\
mg         & 8.67        & 8.90        & 7.64         & 7.24        & \textbf{9.55}  & 0.88        \\
mk         & 6.14        & 6.37        & 5.45         & 4.87        & \textbf{7.67}  & 1.52        \\
my         & 1.51        & 1.74        & 1.31         & 1.09        & \textbf{2.15}  & 0.64        \\
nl         & 6.21        & 6.42        & 5.28         & 5.05        & \textbf{7.57}  & 1.35        \\
or         & 0.40        & 0.43        & 0.29         & 0.29        & \textbf{0.48}  & 0.09        \\
pl         & 6.57        & 6.77        & 5.39         & 5.28        & \textbf{7.86}  & 1.29        \\
pt         & 6.95        & 7.18        & 5.86         & 5.46        & \textbf{8.71}  & 1.77        \\
ro         & 3.33        & 3.53        & 2.78         & 2.58        & \textbf{4.33}  & 1.00        \\
ru         & 8.16        & 8.34        & 7.00         & 6.81        & \textbf{9.61}  & 1.45        \\
sq         & 4.03        & 4.31        & 3.54         & 3.25        & \textbf{5.26}  & 1.22        \\
sr         & 5.39        & 5.62        & 4.38         & 4.22        & \textbf{6.63}  & 1.24        \\
sv         & 4.88        & 5.06        & 3.93         & 3.75        & \textbf{6.07}  & 1.19        \\
sw         & 5.22        & 5.47        & 4.43         & 3.98        & \textbf{6.17}  & 0.95        \\
tr         & 3.54        & 3.68        & 2.99         & 2.66        & \textbf{4.29}  & 0.74        \\
ur         & 3.61        & 3.76        & 2.98         & 2.77        & \textbf{4.48}  & 0.87        \\
zhs        & 6.22        & 6.43        & 5.10         & 4.90        & \textbf{7.19}  & 0.97        \\
zht        & 6.41        & 6.58        & 5.38         & 5.20        & \textbf{7.39}  & 0.98        \\ \midrule
Average    & 5.19        & 5.39        & 4.40         & 4.17        & \textbf{6.25}  & 1.07        \\ \bottomrule
\end{tabular}%
}
\end{table}

In order to gather further insights on the translation capabilities of \projectNameIII and to better visualize in which cases does \projectNameIII achieve performance improvements with respect to other baselines, we conduct further analysis using language similarity and corpora availability lenses. 

\begin{figure}[h]
    \centering
    \caption{BLEU score obtained for each of the bilingual translations tasks using several translation models, grouped by the number of sentence pairs available for the task. Results for tasks with more than 100k sentences are omitted to ease visualization.}
        \includegraphics[width=0.95\textwidth]{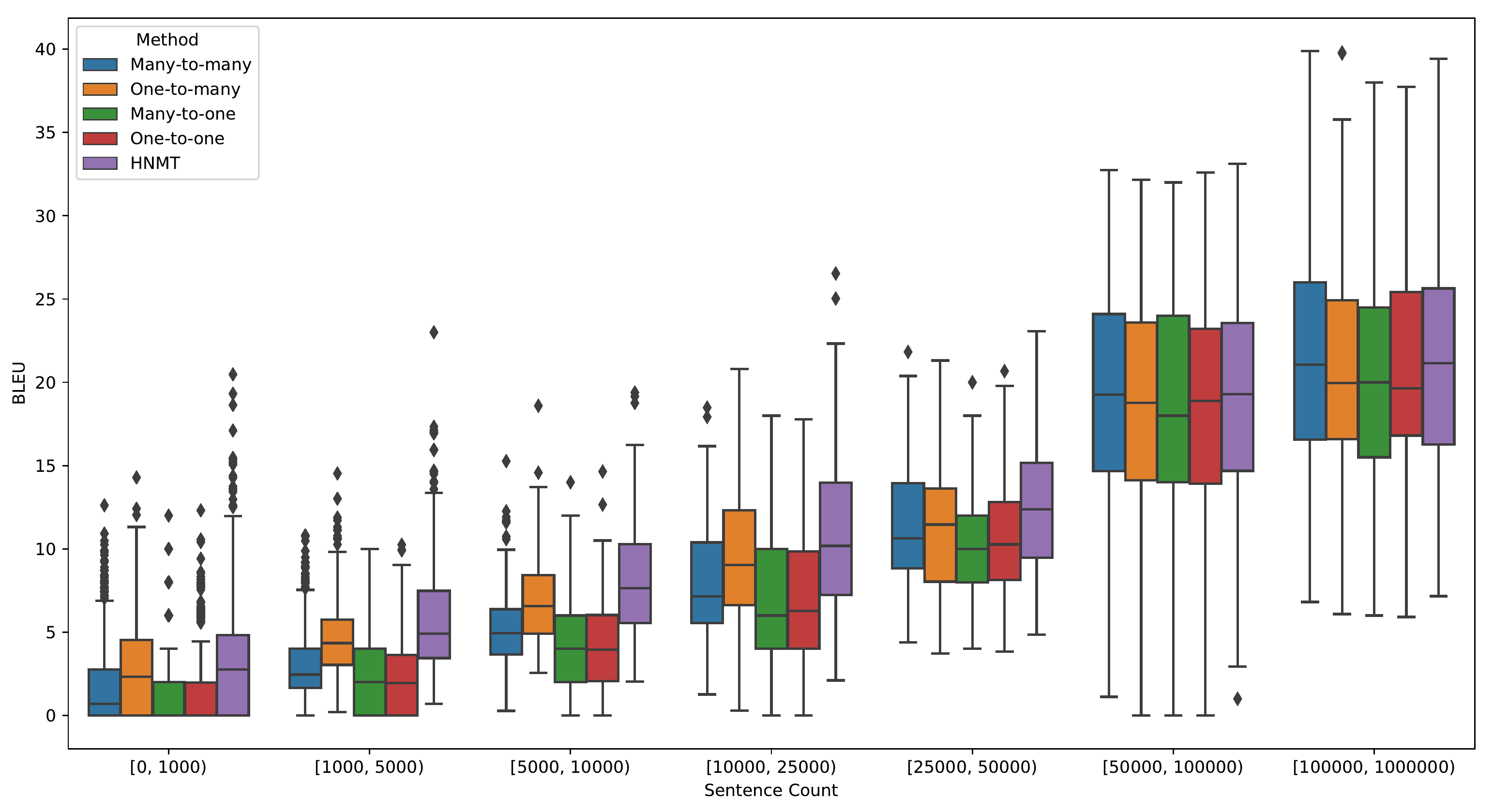}   
     \label{fig:obj3:boxplotSentenceCountAll}
\end{figure}

We first explore model performance when corpora of different sizes is used for training purposes. To do so, we grouped each of the language pair tasks into seven different groups based on the amount of parallel sentence available. As depicted in   Figure~\ref{fig:obj3:boxplotSentenceCountAll}, corpora availability is a determinant factor for translation. The  pattern we devised in Figure~\ref{fig:obj3:scatterPlotSentenceCountM1} is once again visible in Figure~\ref{fig:obj3:boxplotSentenceCountAll}, the more sentences available for a task the better is its performance. However, differences with respect to the baseline are what make a model stand out in this case. Excluding the cases with high amount of corpora, where improvement is hardly possible from a language transfer perspective, we see that  \projectNameIII is the model that achieves the most improvement with respect to the bilingual baseline, followed by the one-to-many model. This behavior denotes that \projectNameIII is indeed capable of improving the performance of machine translation in cases where resources are not abundant.

\begin{figure}[h]
    \centering
    \caption{BLEU score obtained for the models considered in our analysis, grouped by the similarity between the source and target language in the task.}
        \includegraphics[width=0.95\textwidth]{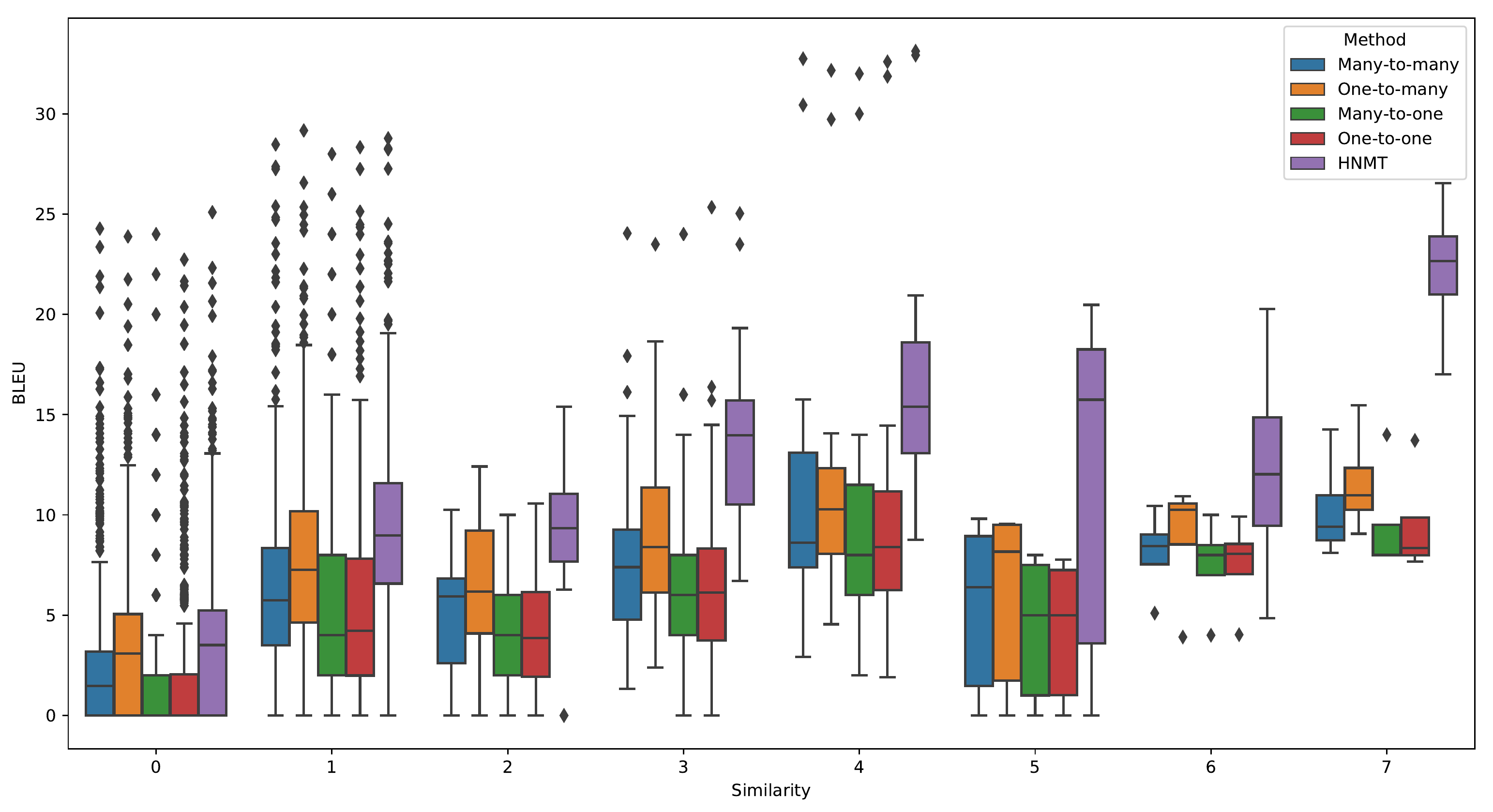}   
   
     \label{fig:obj3:boxplotSimilarityAll}
    
\end{figure}

We are also interested in exploring the effect language similarity has on translation, which is why we examine model performance for languages pairs with different degrees of similarity between them. Results from this experiment are summarized in Figure~\ref{fig:obj3:boxplotSimilarityAll}. 

In general, we observe similar patterns to the ones we previously described: BLEU scores computed for machine translation task are higher for languages that are  similar. This pattern occurs regardless of the language, however, it is considerably more pronounced in the case of \projectNameIII, leading to a higher  improvement with respect to the bilingual baseline the more similar the languages are. We also notice from Figure~\ref{fig:obj3:boxplotSimilarityAll} that none of the other multilingual machine translation models takes advantage of this behavior, maintaining a similar difference with respect to the baseline regardless of the degree of similarity between the languages in pairs considered from analysis. These results serve as indication that the hierarchical organization used in \projectNameIII is indeed useful for explicitly taking advantage of similarities across languages, validating our premises for the design of \projectNameIII.





\section{Conclusion and Future Work}
In this paper, we presented \projectNameIII, an hierarchical framework for machine translation that can be applied to any multilingual neural machine translation strategy, for achieving a higher degree of transfer learning across languages. We conducted several experiments using 758 language pairs including languages with varied resource availability and similarity. Our empirical analysis reveals that highest improvements take place when the languages are typologically related and aligned corpora is not abundant, achieving an improvement of about 5 BLEU points in specific cases. These results validate our premise that machine translation for low-resource languages can be enhanced by means of language transfer if an appropriate organization of languages is used, such as the one we utilize as part of \projectNameIII. As a natural part of its encoding-decoding process for translation, \projectNameIII generates a language-agnostic vector representation of sentences. While we did not evaluated the quality of this by-product of our work, given that it was out of scope, exploratory examinations lead us to believe that these language-agnostic representations could be leveraged for supporting a multilingual applications in related text processing areas.

We are aware of some limitations of this work. First,  even if the strategy is shown to improve low-resource scenarios where the source and target language are typologically related, this effect is not as prominent when the languages are different from each other. Consequently, the applicability of  \projectNameIII for isolated languages such as Basque is limited. Second,  given the high amount of tasks and languages considered, the size of the machine translation models we used for experimentation is small compared to current state-of-the-art systems. For example, we set $|h|=512$ and $N=5$, when current strategies use $|h|=1024$ and $N=8$. In the future, we plan on leveraging other types of signals, such as the use of sub-word embeddings, for enabling further language transfer. Additionally, we will extend our empirical analysis to explore the performance effect of using larger and more varied machine translation models, such as Convolutional Neural Networks or Transformers.

\bibliographystyle{plainnat}  
\bibliography{IonReferences}  


\end{document}